%% file: main.tex
\documentclass{article}

\usepackage[preprint]{corl_2021} 

\input{packages}

\title{Group-based Motion Prediction for Navigation in Crowded Environments}

\author{
  Allan Wang\\
  Robotics Institute\\
  Carnegie Mellon University\\
  \And
  Christoforos Mavrogiannis\\
  Paul G. Allen School of Computer Science \& Engineering\\
  University of Washington\\
  \And
  Aaron Steinfeld\\
  Robotics Institute\\
  Carnegie Mellon University\\
}

\begin{document}
\maketitle

\input{abstract}

\input{sections/introduction}
\input{sections/related-work}

\input{sections/problem-statement}
\input{sections/group}

\input{sections/planning}
\input{sections/evaluation}
\input{sections/discussion}

\input{sections/acknowledgment}

\bibliography{references}

\appendix
\input{sections/appendix.tex}

\end{document}

%% file: packages.tex
\usepackage{amssymb,amsmath}
\usepackage{booktabs}
\usepackage{caption}
\usepackage{subcaption}
\usepackage{graphicx}
\usepackage{graphics}

\newcommand{\figref}[1]{Fig.~\ref{#1}}
\newcommand{\tabref}[1]{Table~\ref{#1}}

\usepackage{multicol,multirow}
\usepackage{verbatimbox}

\usepackage{hyperref}
\hypersetup{bookmarksopen,bookmarksnumbered,
pdfpagemode=UseOutlines,
colorlinks=true,
linkcolor=blue,
anchorcolor=blue,
citecolor=blue,
filecolor=blue,
menucolor=blue,
urlcolor=blue
}

\usepackage{xcolor}

\newcommand{\xxnote}[3]{}
\ifx\hidenotes\undefined
  \usepackage{color}
  \renewcommand{\xxnote}[3]{\color{#2}{#1: #3}}
\fi

%% file: abstract.tex
\begin{abstract}
We focus on the problem of planning the motion of a robot in a dynamic multiagent environment such as a pedestrian scene. Enabling the robot to navigate safely and in a socially compliant fashion in such scenes requires a representation that accounts for the unfolding multiagent dynamics. Existing approaches to this problem tend to employ microscopic models of motion prediction that reason about the individual behavior of other agents. While such models may achieve high tracking accuracy in trajectory prediction benchmarks, they often lack an understanding of the group structures unfolding in crowded scenes. Inspired by the Gestalt theory from psychology, we build a Model Predictive Control framework (G-MPC) that leverages group-based prediction for robot motion planning. We conduct an extensive simulation study involving a series of challenging navigation tasks in scenes extracted from two real-world pedestrian datasets. We illustrate that G-MPC enables a robot to achieve statistically significantly higher safety and lower number of group intrusions than a series of baselines featuring individual pedestrian motion prediction models. Finally, we show that G-MPC can handle noisy lidar-scan estimates without significant performance losses.
\end{abstract}

%% file: sections/introduction.tex
\section{Introduction}

Over the past three decades, there has been a vivid interest in the area of robot navigation in pedestrian environments~\citep{minerva-robot,kruse13-navSurvey,kretzschmar_ijrr16,trautmanijrr,Mavrogiannis19}. Planning robot motion in such environments can be challenging due to the lack of rules regulating traffic, the close proximity of agents and the complex emerging multiagent interactions. Further, accounting for human safety and comfort as well as robot efficiency add to the complexity of the problem. 

To address such specifications, a common~\citep{kretzschmar_ijrr16,trautmanijrr,learning-social-robot,Kim2016,Everett18_IROS} paradigm involves the integration of a behavior prediction model into a planning mechanism. Recent models tend to predict the individual interactions among agents to enable the robot to determine collision-free candidate paths~\citep{kretzschmar_ijrr16,trautmanijrr,MavBluKne_IROS2017}. While this paradigm is well-motivated, it tends to ignore the structure of interaction in such environments. Often, the motion of pedestrians is coupled as a result of social grouping. Further, the motion of multiple agents can often be \emph{effectively} grouped as a result of similarity in motion characteristics. Lacking a mechanism for understanding the emergence of this structure, the robot motion generation mechanism may yield unsafe or uncomfortable paths for human bystanders, often violating the space of social groups.

Motivated by such observations, we draw inspiration from human navigation to propose the use of group-based prediction for planning in crowd navigation domains. We argue that humans do not employ detailed individual trajectory prediction mechanisms. In fact, our motion prediction capabilities are short-term and do not scale with the number of agents. We do however employ effective grouping techniques that enable us to discover safe and efficient paths among motions of crowd networks. This anecdotal observation is aligned with gestalt theory from psychology~\citep{koffka35} which suggests that organisms tend to perceive and process \emph{formations of entities}, rather than individual components. Such techniques have recently led to advances in computer vision~\citep{Desolneux07} and computational photography~\citep{vazquez11}. Similarly, we envision that a robot could reason the formation of effective groups in a crowded environment and react to their motion as an effective way to navigate safely.

In this paper, we propose a group-based representation coupled with a prediction model based on the group-space approximation model of \citet{wang-split-merge}. This model groups a crowd into sets of agents with similar motion characteristics and draws geometric enclosures around them, given observation of their states. The prediction module then predicts future states of these enclosures. We conduct an extensive empirical evaluation over 5 different human datasets~\citep{ETH,UCY}, each with a flow following and a crossing scenario. We further conduct a same set of evaluations with agents powered by ORCA~\citep{ORCA} that share the start and end locations in the datasets. Last but not least, we conducted evaluation given inputs in the form of simulated laser scans, from which pedestrians are only partially observable or even completely occluded. We compare the performance of our group-based formulation against three individual reasoning baselines: a) a reactive baseline with no prediction; b) a constant velocity prediction baseline; c) one based on individual S-GAN trajectory predictions~\citep{socialGan}. We present statistically significant evidence suggesting that agents powered by our formulation produce safer and more socially compliant behavior and are potentially able to handle imperfect state estimates.

\begin{figure}
\centering
\includegraphics[width=0.9\linewidth]{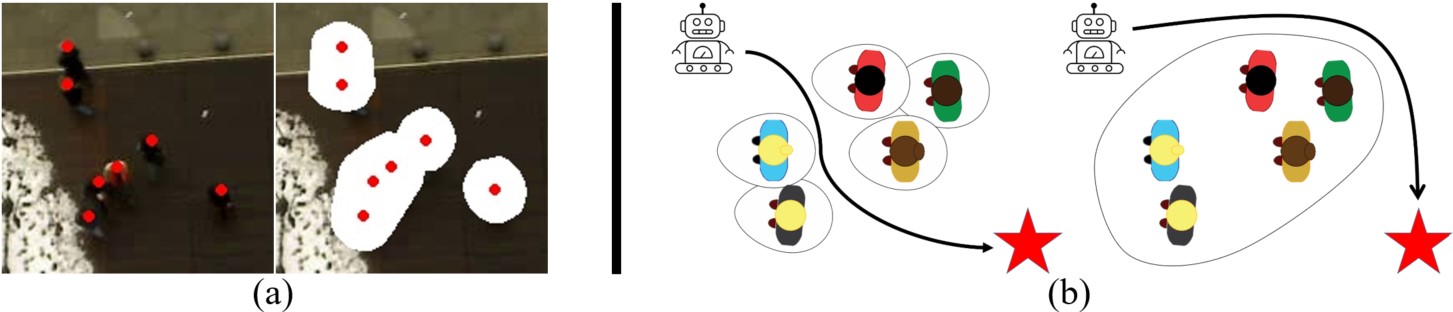}
\caption{Based on a representation of social grouping~\citep{wang-split-merge}, we build a group behavior prediction model to empower a robot to perform safe and socially compliant navigation in crowded spaces. (a) Example of our representation overlayed on top of a scene from a real-world dataset~\citep{ETH}. (b) A model predictive controller equipped with our prediction model is able to navigate around the group socially (right) as opposed to the baseline that cuts through the group (left). }
\label{fig:personal_space}
\vspace{-1em}
\end{figure}

%% file: sections/related-work.tex
\section{Related Work}

Over the recent years, a considerable amount of research has been placed to the problem of robot navigation in crowded pedestrian environments~\citep{kretzschmar_ijrr16,trautmanijrr,Everett18_IROS,Mavrogiannis18,chen2019crowd,zhi2020anticipatory,Somani2013,Cai2021,Kiss2021}. Such environments often comprise groups of pedestrians, navigating as coherent entities. 
%
\citet{Sochman-2011} suggests that $50$-$70\%$ of pedestrians walk in groups. Many works exist in group detection. One popular area in such domain is static group detection, often leveraging F-formation theories~\citep{Kendon-1990-Fformation}. However, dynamic groups often dominate pedestrian-rich environments and exhibit different spatial behavior~\cite{Yang-2019-groupnav}. Among dynamic group detection, one common approach is to treat grouping as a probabilistic process where groups are a reflection of close probabilistic association of pedestrian trajectories~\citep{bazzani-2012, Chang-2011-group, Gennari-2004, Pellegrini-2010-group, Zanotto-2012}. Others use graph models to build inter-pedestrian relationships with strong graphical connections indicating groups~\citep{chamveha-2013, Khan-2015}. The social force model~\citep{social-force} also inspires \citet{Mazzon-2013, Sochman-2011} to develop features that indicate groups. Clustering is another common group of technique to group pedestrians with similar features into groups~\citep{solera-2016, Ge-2012-group, taylor-2020-group, Chatterjee1}. For our formulation, it is sufficient to employ a simple clustering-based grouping method proposed by \citet{Chatterjee1}. Other grouping methods will simply result in different group membership assignments.

Applications on groups often focus on a specific behavior aspect. For example, one focus in this area is how a robot should behave as part of the group formation~\citep{Cuntoor-2012}. On dyad groups involving a single human and a robot, some researchers examined socially appropriate following behavior~\citep{Gockley-2007, Granata-2012, Jung-2012, Zender-2007} and guiding behavior~\citep{nanavati-dyad, Feil-Seifer-2011, Pandey-2009}. In works that do not include robots as part of pedestrian groups, some research teams studied how a robot should guide a group of pedestrians~\citep{Garrell-2010, Shiomi-2007, Martinez-Garcia-2005}. From navigation perspective, Yang and Peters~\citep{Yang-2019-groupnav} leverage groups as obstacles, but their group space only involves occasional O-space modeling from F-formation theories. Without the engineered occurrence of O-space, their representation reduces to one of our baselines. Katyal et al.~\citep{Katyal-2020-groupnavRL} introduce an additional cost term that leverages robot's distance to the closest group in a reinforcement learning framework. They model groups using convex hulls directly generated from pedestrian coordinates instead of taking personal spaces into consideration. 
In our work, we additionally explore the capabilities of groups in handling imperfect sensor inputs. While our focus is on analysing the benefits of groups, our group based formulation can be easily incorporated into the work of~\citet{Katyal-2020-groupnavRL}'s framework.

%% file: sections/problem-statement.tex
\section{Problem Statement}

Consider a robot navigating in a workspace $\mathcal{W}\subseteq\mathbb{R}^2$ amongst $n$ other dynamic agents. Denote by $s\in\mathcal{W}$ the state of the robot and by $s^i\in\mathcal{W}$ the state of agent $i\in\mathcal{N} = \{1,\dots, n\}$. The robot is navigating from a state $s_{0}$ towards a destination $s_{T}$ by executing a policy $\pi:\mathcal{W}^{n+1}\times\mathcal{U}\to\mathcal{U}$ that maps the assumed fully observable world state $\boldsymbol{S} = s \cup_{i = 1:n} s^i$ to a control action $u\in\mathcal{U}$, drawn from a space of controls $\mathcal{U}\subseteq\mathbb{R}^2$. We assume that the robot is not aware of agents' destinations $s^{i}_T$ or policies $\pi_i:\mathcal{W}^{n+1}\times\mathcal{U}^i\to\mathcal{U}^i$, $i\in\mathcal{N}$. In this paper, our goal is to design a policy $\pi$ that enables the robot to navigate from $s_0$ to $s_T$ safely in a socially compliant fashion.

%% file: sections/group.tex
\section{Group-based Prediction}\label{sec:prediction}

We introduce a group representation building on prior work~\citep{wang-split-merge} and a model for group-based prediction that is amenable for use in decentralized multiagent navigation.

\subsection{Group Representation}

Define as $\theta^i\in [0, 2\pi)$ the orientation of agent $i\in \mathcal{N}$ which is assumed to be aligned with the direction of its velocity $u^i$, extracted via finite differencing of its position over a timestep $dt$ and denote by $v^i = ||u^i||\in \mathbb{R}^{+}$ its speed. We define an augmented state for agent $i$ as $q^i = (s^i, \theta^i, v^i)$.

We treat a social group as a set of agents who are in close proximity and share similar motion characteristics. Assume that a set of $J$ groups, $\mathcal{J} = \{1,\dots, J\}$ navigate in a scene. Define by $g^i\in\mathcal{J}$ a label indicating the group membership of agent $i$. We then define a group $j\in\mathcal{J}$ as a set $G^j = \{i \in \mathcal{N} \mid g^i = j \}$ and collect the set of all groups in a scene into a set $\boldsymbol{G} = \{G^j\mid j\in\mathcal{J}\}$.

\label{sec:group-membership}\textbf{Extracting Group Membership.} We define the combined augmented state of all agents as $\boldsymbol{q} = \cup_{i=1:n}q^i$. To obtain group memberships for a set of agents $\mathcal{N}$, we apply the Density-Based Spatial Clustering of Applications with Noise algorithm (DBSCAN)~\citep{dbscan} on agent states:
\begin{equation}
\label{eq:dbscan}
    \boldsymbol{G}\longleftarrow \texttt{DBSCAN}(\boldsymbol{q} \mid \epsilon_s, \epsilon_\theta, \epsilon_v) 
\end{equation}
Where $\epsilon_s, \epsilon_\theta, \epsilon_v$ are respectively threshold values on agent distances, orientation and speeds. 




\label{sec:groupspacerep}\textbf{Extracting the Social Group Space.} For each group $G^j$, $j\in\mathcal{J}$, we define a \emph{social group space} as a geometric enclosure $\mathcal{G}^j$ around agents of the group. For each agent $i\in G^j$, we define a personal space $\mathcal{P}^i$ as a two-dimensional asymmetric Gaussian based on the model introduced by \citet{kirby-2010}. Refer to Appendix A for detailed descriptions.


Given the personal spaces $\mathcal{P}^i$, $i\in G^j$, of all agents in a group $j$, we extract the social group space of the whole group as a convex hull:
\begin{equation}
    \mathcal{G}^j=\mbox{Convexhull}(\{\mathcal{P}^i\mid i\in G^j\})\mbox{.}
\end{equation}
The shape described by $\mathcal{G}^j$ represents an obstacle space representation of a group containing agents in close proximity with similar motion characteristics. For convenience, let us collect the spaces of all groups in a scene into a set $\boldsymbol{\mathcal{G}} = \{\mathcal{G}^j\mid j\in \mathcal{J}\}$.

\subsection{Group Space Prediction Oracle}\label{sec:prediction-oracle}

Based on the group-space representation of Sec.~\ref{sec:groupspacerep}, we describe a prediction oracle that outputs an estimate of the future spaces occupied by a set of groups $\boldsymbol{\mathcal{G}}_{t:t_f}$ up to a time $t_f = t+ f$, where $f$ is a future horizon given a past sequence of group spaces $\boldsymbol{\mathcal{G}}_{t_h:t}$ from time $t_h = t -h$ where $h$ is a window of past observations:
\begin{equation}
    \boldsymbol{\mathcal{G}}_{t:t_f} \leftarrow \mathcal{O}(\boldsymbol{\mathcal{G}}_{t_h:t}) = \cup_{j=1:J}\mathcal{O}_j(\mathcal{G}^j_{t_h:t})\mbox{,}\label{eq:oracledef}
\end{equation}
where $\mathcal{O}_j$ is a model generating a group space prediction for group $G^j$. Refer to Appendix B for detailed description of partial input handling.



We implement the oracle $\mathcal{O}_j$ of eq.~\eqref{eq:oracledef} using a simple encoder-decoder network. The encoder follows the 3D convolutional architecture in \cite{Tran-2015} whereas the decoder mirrors the model layout of the encoder. The encoder-decoder network takes as input a sequence\footnote{The oracle input sequence is first converted into image-space coordinates using the homography matrix of the scene. We also preprocess inputs to have normalized scale and group positions. The autoencoder output is converted back into Cartesian coordinates using the inverse homography transform.} $\mathcal{G}_{t_h:t}$ and outputs a sequence $\mathcal{G}_{t+1:t_f}$ which we pass through a sigmoid layer. We supervise the encoder-decoder network's output using the binary cross entropy loss.


We verified the effectiveness of our encoder-decoder network on the 5 scenes of our experiments by conducting a cross-validation comparison against a baseline. The baseline predicts the future shapes by linearly translating the last social group shape using its geometric center velocity. We use Intersection over Union (IoU) as our metric. Between the ground truths and the predictions, this metric divides the number of overlapped pixels by the number of pixels occupied by either one of them. As shown in Table \ref{tab:autoencoder_perf}, our encoder-decoder network outperforms the baseline. 

\begin{table}
    \centering
        \caption{Prediction Model Performance}
    \resizebox{.8\columnwidth}{!}{%
    \begin{tabular}{c|c|c|c|c|c|c}
        \toprule 
        & Metric & ETH & HOTEL & ZARA1 & ZARA2 & UNIV \\
        \midrule
        \multirow{2}{*}{Baseline} & $\mbox{mIoU}$ (\%) & 83.52 & 90.37 & 88.04 & 89.30 & 85.32 \\ 
        & $\mbox{fIoU}$ (\%) & 76.32 & 85.38 & 82.14 & 83.88 & 77.24\\
        \midrule
        \multirow{2}{*}{Encoder-decoder Network} & $\mbox{mIoU}$ (\%) & 86.66 & 92.10 & 89.97 & 90.94 & 87.52 \\ 
        & $\mbox{fIoU}$ (\%) & 78.64 & 86.83 & 83.77 & 85.09 & 78.55\\
        \bottomrule
    \end{tabular}
    }
    \label{tab:autoencoder_perf}
    \vspace{-1em}
\end{table}

%% file: sections/planning.tex
\section{Model Predictive Control with Group-based Prediction}\label{sec:mpc}

We describe G-MPC, a model predictive control (MPC) framework for navigation in multiagent environments that leverages the group-based prediction oracle of Sec.~\ref{sec:prediction}.


We describe our group-prediction informed MPC, or G-MPC. At planning time $t$, given a (possibly partial) augmented world state history $\boldsymbol{Q}_{t_{\hat{h}} : t}$, we first extract a sequence of group spaces $\boldsymbol{\mathcal{G}}_{t_{h}:t}$ based on the method of Sec.~\ref{sec:groupspacerep}. Given these, the robot computes an optimal control trajectory $\boldsymbol{u}^* = u^*_{1:K}$ of length $K$ by solving the following optimization problem:
\begin{align}
\left(\boldsymbol{s}^*, \boldsymbol{u}^{*}\right) =&  \arg\min_{u_{1:K}} \sum_{k=1:}^{K} \gamma^{k}J(s_{k+1}, \boldsymbol{\mathcal{G}}_{k+1}, s_T)\label{eq:cost}\\
 s.t.\: 
         & \boldsymbol{\mathcal{G}}_{2-h:1} \leftarrow \boldsymbol{\mathcal{G}}_{t_{h}:t}\label{eq:initgroups}\\
         & s_1 \leftarrow s_t\label{eq:initrobot}\\
         & \boldsymbol{\mathcal{G}}_{k+1:k_{f}} = \mathcal{O}(\boldsymbol{\mathcal{G}}_{k_{h}:k})\label{eq:updategroup}\\
         & u_k\in\mathcal{U}\\
         & s_{k+1} = s_k + u_k \cdot dt\label{eq:statetransition}
   \mbox{,}
\end{align}
where $\gamma$ is the discount factor and $J$ represents a cost function, eq.~\eqref{eq:initgroups} initializes the group space history ($k = 2-h$ is the timestep displaced a horizon $h$ in the past from the first MPC-internal timestep $k=1$), eq.~\eqref{eq:initrobot} initializes the robot state to the current robot state $s_t$, eq.~\eqref{eq:updategroup} is an update rule recursively generating a predicted future group sequence up to timestep $k_f = k+f$ given history from time $k_h = k - h$ up to time $k$, $\mathcal{O}$ represents a group-space prediction oracle based on Sec.~\ref{sec:prediction}, and eq.~\eqref{eq:statetransition} is the robot state transition assuming a fixed time parametrization of step size $dt$.

We employ a weighted sum of costs $J_g$ and $J_d$, penalizing respectively distance to the robot's goal and proximity to groups:
\begin{equation}
    J(s_{k}, \boldsymbol{\mathcal{G}}_{k}, s_T) = \lambda J_{g}(s_{k}, s_T) +(1-\lambda)J_{d}(s_{k}, \boldsymbol{\mathcal{G}}_k)
    \label{eq:begin_detail_mpc}\mbox{,}
\end{equation}
where $\lambda$ is a weight representing the balance between the two costs and
\begin{equation}
    J_{g}(s_{k}) =\begin{cases}
    0, & \mbox{if}\: s_{k}\in \boldsymbol{\mathcal{G}}_{k}\\
    ||s_{k-1} - s_T||, & \mbox{else,}
    \end{cases}
\end{equation}
penalizes a rollout according to the distance of the last collision-free waypoint to the robot's goal. Further, we define $J_d$ as:
\begin{equation}
    J_{d}(s_{k}, \boldsymbol{\mathcal{G}}_k) = \exp(-\mathcal{D}\left(s_{k+1}, \boldsymbol{\mathcal{G}}_k)\right)\mbox{,}
\end{equation}
where
\begin{equation}
    \mathcal{D}(s_{k}, \boldsymbol{\mathcal{G}}_k) = \begin{cases}
                \min_{j\in \mathcal{J}} D\left(s_k - \mathcal{G}_k^j\right), \: 
                & \mbox{if}\: s_{k}\notin \mathcal{G}^j_{k}\\
                -\min_{j\in \mathcal{J}} D\left(s_k - \mathcal{G}_k^j\right), \:
                &\mbox{else}\mbox{,}
               \end{cases}
    \label{eq:end_detail_mpc}
\end{equation}
where $D(s_k - \mathcal{G}_k^j)$ returns the minimum distance between the robot state and the space occupied by group $j$ at time $k$. Using $D$, function $\mathcal{D}$ computes the minimum distance to any group for a given time. In most cases, the robot lies outside of groups, i.e., $s_{k}\notin \mathcal{G}^j_{k}$ --therefore, the cost $J_d$ tries to maximize the distance $\mathcal{D}$. Sometimes, the robot might end up entering the group space $\boldsymbol{\mathcal{G}}$ --in those cases, $J_d$ tries to minimize $\mathcal{D}$, to steer the robot towards the direction of quickest escape from the group. In case that the robot is inside a group to begin with, we shrink the group sizes in Sec.~\ref{sec:groupspacerep} until the robot is outside the groups again.








To solve eq.~\eqref{eq:cost}, we search over a finite set $\boldsymbol{\mathcal{U}}$ of control trajectories of horizon $K$. With the assumption that the robot is holonomic and is not under any kinematic constraints, we use a set of $R$ control rollouts $\boldsymbol{\mathcal{U}} = \{\boldsymbol{u}^1,...,\boldsymbol{u}^R\}$ with three levels of tangential speeds and a set of turning speed, i.e.,
\begin{equation}
    \label{eq:rollout}
    u_{1:K}^r = (v\cos\psi, v\sin\psi, \omega),\: \psi=\frac{2\pi r}{R},
    v\in\left\{\frac{1}{3}v_{max}, \frac{2}{3}v_{max}, v_{max}\right\},
    \omega\in\left\{0, \pm\frac{\pi}{2}\right\}
\end{equation}
To ensure compatibility between our group-based prediction model and our MPC formulation, we set the control rollout time horizon to be the prediction model's prediction horizon, or $K=f$.











%% file: sections/evaluation.tex
\section{Evaluation}\label{sec:evaluation}

We evaluate our framework through a simulation study in which the robot performs a navigation task (a transition between two points) within a crowds of dynamic agents in a set of scenes. 

\subsection{Experimental Setup}

\begin{figure}[t]
\centering
\includegraphics[width=\linewidth]{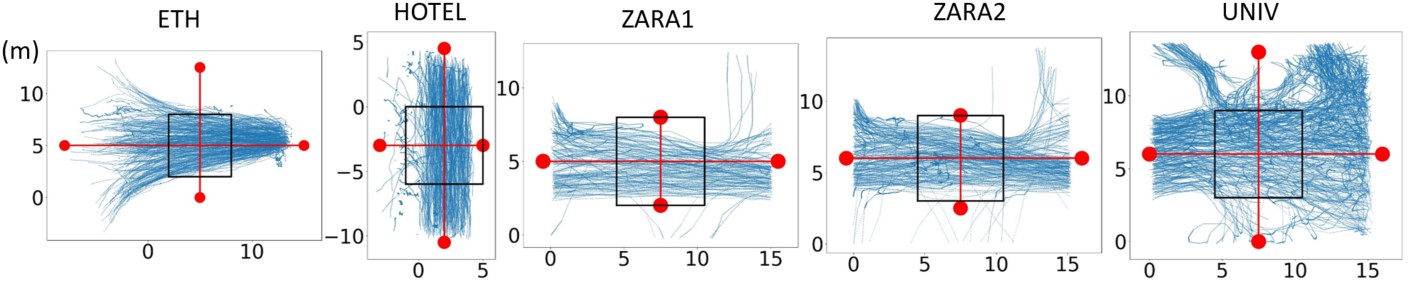}
\caption{Trajectories of all pedestrians in the datasets. The red dots represent the task start and end locations. The red lines represent the task paths. The black box represents the test region to check for non-trivial tasks.}
\label{fig:test_cases}
\vspace{-2em}
\end{figure}

We consider a set of realistic pedestrian scenes, drawn from the ETH~\citep{ETH} (ETH and HOTEL scenes) and UCY~\citep{UCY} (ZARA1, ZARA2 and UNIVERSITY scenes) datasets, which often serve as benchmarking testbeds in the motion prediction and social navigation literature~\citep{socialGan, socialLstm,  SRLSTM,dynamic-channel,biswas2021socnavbench}. In each scene, we define two navigation tasks (see~\figref{fig:test_cases}): \textit{Flow}: in which the robot navigates along the crowd flow and \textit{Cross} in which the robot intersects vertically with the traffic flow. For each task, we generate a set of trials by segmenting the scene recording into blocks involving challenging interactions. We define a challenging interaction to be a segment involving at least $5$ pedestrians inside the test region drawn in black in~\figref{fig:test_cases}. This process provided us with a distribution of trials as shown in table~\tabref{tab:trialsperscene}. Across all trials, we keep the robot's maximum speed at $1.75 m/s$.

\begin{table}
    \centering
        \caption{Number of trials per task and scene.}
            \resizebox{.6\columnwidth}{!}{%
    \begin{tabular}{c|c|c|c|c|c}
        \toprule 
        Task & ETH & HOTEL & ZARA1 & ZARA2 & UNIV \\
        \midrule
        Flow & 58 & 43 & 25 & 127 & 106\\
        Cross & 58 & 44 & 28 & 129 & 114\\
        \bottomrule
    \end{tabular}
    }
    \label{tab:trialsperscene}
    \vspace{-2em}
\end{table}

We consider two experimental conditions: an \textit{Offline} and an \textit{Online} one. In the \textit{Offline} one, the robot navigates among a crowd moving according to a recording of a human crowd. Under this condition, pedestrians act as dynamic obstacles that do not react to the robot, a situation which could arise in cases where robots are of shorter size and could thus be easily missed by navigating pedestrians. In the \textit{Online} one, the robot navigates among a crowd\footnote{For consistency, the agents in the crowd start and end at the same spots as the agents in the recorded crowd from the Offline condition.} moving by running ORCA~\citep{ORCA}, a policy that is frequently used as a simulation engine for benchmarking in the social navigation  literature~\citep{Everett18_IROS,dynamic-channel,mavrogiannis_etal2021-core-challenges}.



To investigate the value of G-MPC, we develop three variants of it. \textbf{group-pred} is a G-MPC in which the encoder-decoder network has a history $h=8$ and a horizon $f = 8$. \textbf{group-nopred} is a variant that features no prediction at all --it just reacts to observed groups at every timesteps and it is equivalent to the framework of~\citet{Yang-2019-groupnav}. Finally, \textbf{laser-group-pred} is identical to \textbf{group-pred} but instead of using ground-truth pose information, it takes as input noisy lidar scan readings. We simulate this by modeling pedestrians as $1m$-diameter circles and lidar scans as rays projecting from the robot. We refer to the spec sheet of a SICK LMS511 2D lidar for simulation parameters. We further inject noise into the readings according to the spec sheet. Under this simulation, pedestrians may only be partially observable or even completely occluded from the robot.

We compare the performance of these policies against a set of MPC variants using mechanisms for individual motion prediction. \textbf{ped-nopred} is a vanilla MPC that reacts to the current states of other agents without making predictions about their future states. \textbf{ped-linear} is a vanilla MPC that estimates future states of agents by propagating agents' current velocities forward. This baseline is motivated by recent work showing that constant-velocity models yield competitive performance in pedestrian motion prediction tasks~\citep{constant-velocity-model}. Finally, \textbf{ped-sgan} is an MPC that uses S-GAN~\citep{socialGan} to extract a sequence of future state predictions for agents based on inputs of their past states. We selected S-GAN because it is a recent highly performing model. To ensure a fair comparison, all the MPC policy variants are integrated with the same MPC controller evaluated at $dt=0.1$.

We measure the performance of the policies with respect to four different metrics: a) \textit{Success rate}, defined as the ratio of successful trials over total number of trials; b) \textit{Comfort}, defined as the ratio of trials in which the robot does not enter any social group space over the total number of trials; c) \textit{Minimum distance to pedestrians}, defined as the smallest distance between the robot and any agent per trial; d) \textit{Path length}, defined as the total distance traversed by the robot in a trial.

To track the performance of G-MPC, we design hypotheses targeting aspects of safety and group space violation which we investigate under both experimental conditions, i.e., offline and online:

\textbf{H1}: To explore the benefits of group based representations alone, we hypothesize that \textbf{group-nopred} is safer than \textbf{ped-nopred} while achieving similar success rates but worse efficiency.

\textbf{H2}: To explore the full benefit of group based formulation, we hypothesize that \textbf{group-pred} is safer than \textbf{ped-linear} and \textbf{ped-sgan} while achieving similar success rates but worse efficiency.

\textbf{H3}: To explore how our formulation handles imperfect inputs, we hypothesize that \textbf{laser-group-pred} achieves similar safety to \textbf{group-pred} while achieving similar success rate and efficiency.

\textbf{H4}: To check that our formulation is socially compliant, we hypothesize that \textbf{group-nopred}, \textbf{group-pred} and \textbf{laser-group-pred} violate agents' group space less often than the baselines.





\subsection{Results}


\begin{figure}
    \centering
    \includegraphics[width=\linewidth]{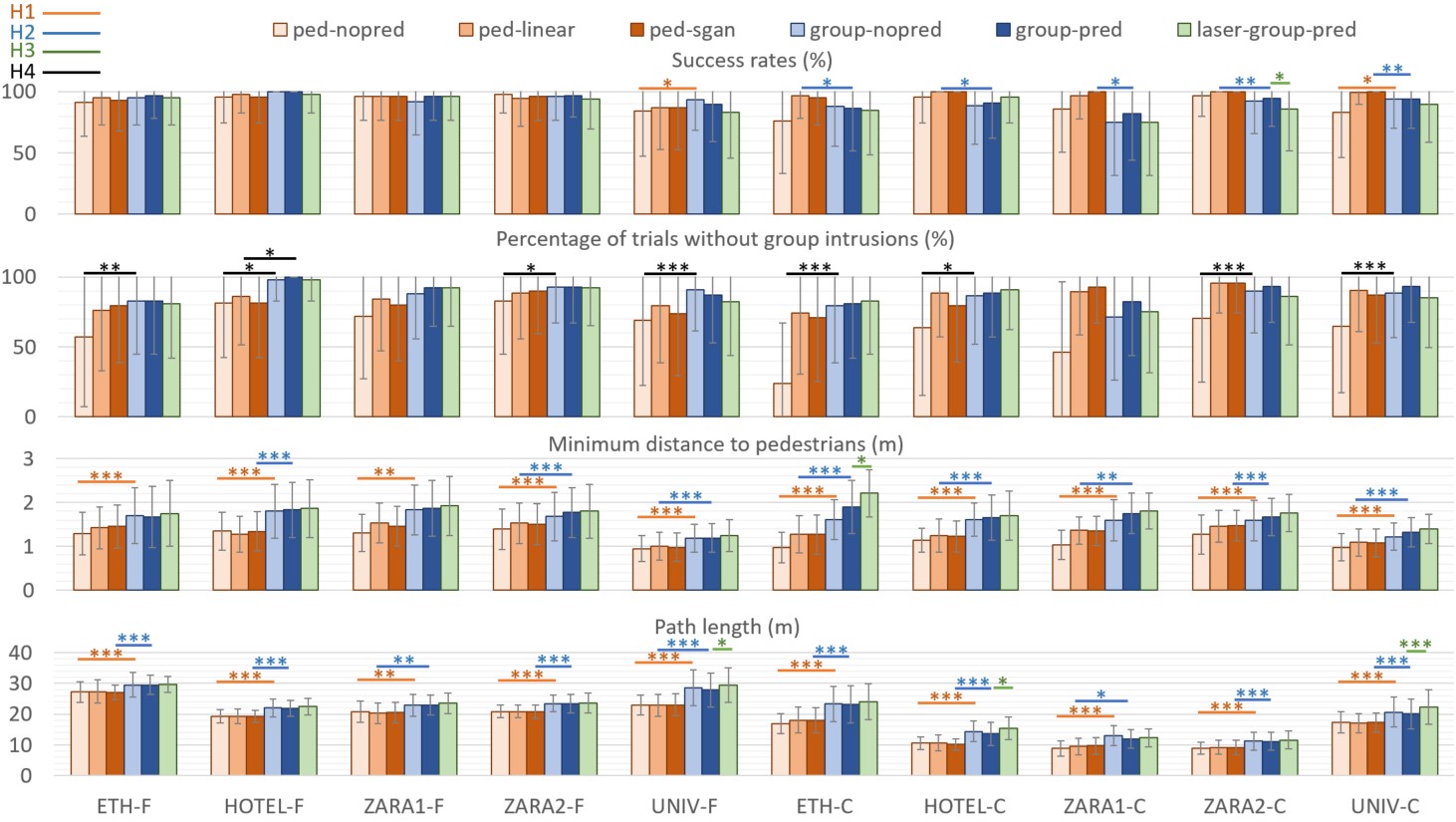}
    \caption{Performance per scene under the \textit{Offline} condition. Horizontal lines indicate statistically significant results corresponding to different to hypotheses.}
    \label{fig:rst_noreact}
    \vspace{-1em}
\end{figure}

\begin{figure}
    \centering
    \includegraphics[width=\linewidth]{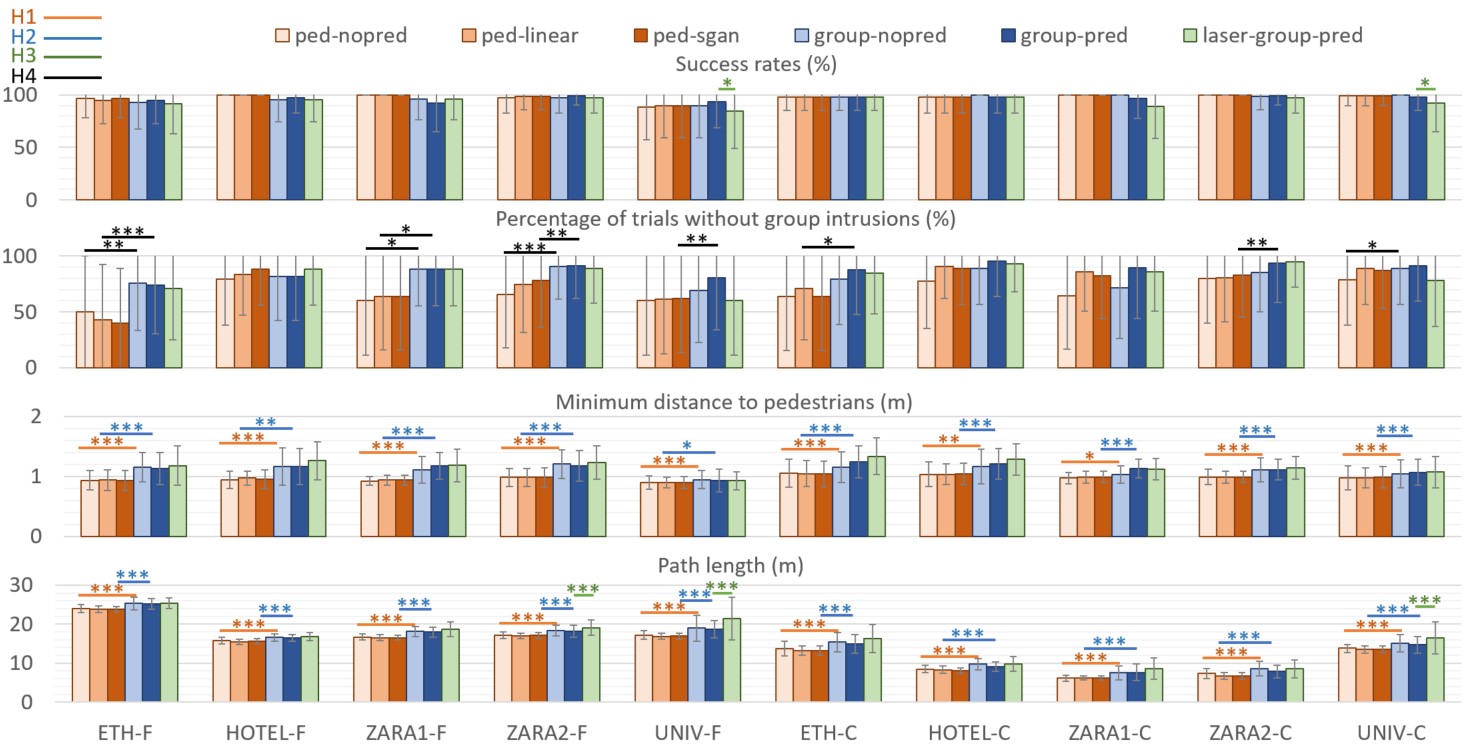}
    \caption{Performance per scene under the \textit{Online} condition (simulated pedestrians powered by ORCA~\cite{ORCA}). Horizontal lines indicate statistically significant results corresponding to different to hypotheses.}
    \label{fig:rst_react}
    \vspace{-1em}
\end{figure}

\begin{figure}
    \centering
    \includegraphics[width=\linewidth]{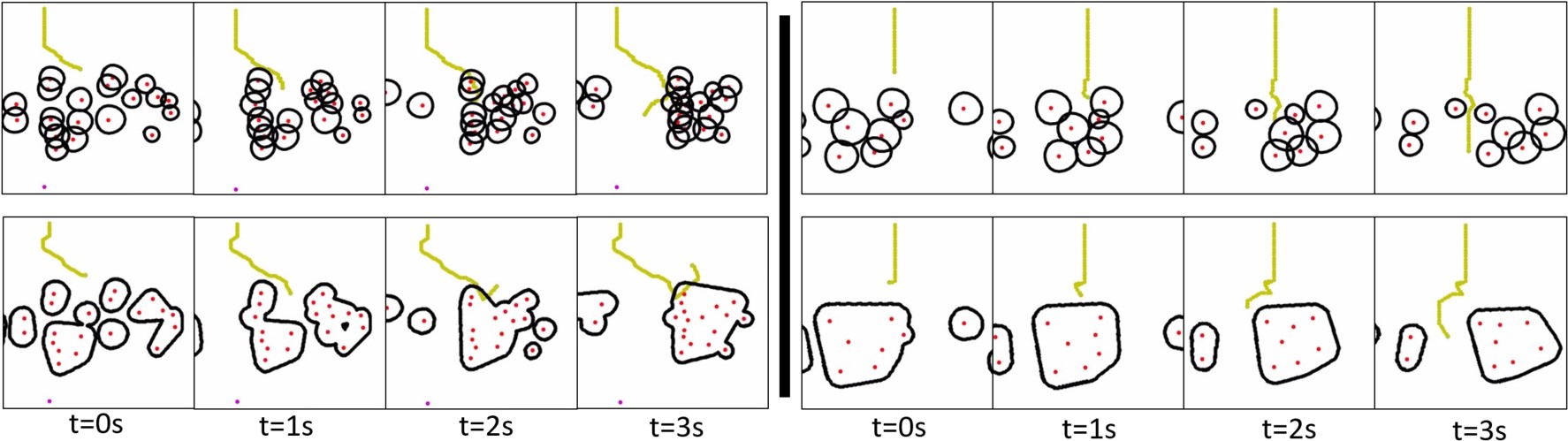}
    \caption{Qualitative performance difference between approaches leveraging pedestrian-based (top) and group-based (bottom) representations. Left: non-reactive agents. Right: reactive agents. In all cases, the robot starts from the top and attempts to navigate downward. The green line shows the robot's traversed trajectory.}
    \label{fig:rst_examples}
    \vspace{-2em}
\end{figure}

\textbf{Quantitative Analysis.} Fig.~\ref{fig:rst_noreact} and Fig.~\ref{fig:rst_react} contain bar charts representing the performance of G-MPC compared with its baselines under Offline and Online settings respectively. Bars indicate means, errorbars indicate standard deviations, ``F" and ``C" are flow and cross scenarios respectively, and the number of asterisks indicates increasing significance levels: $\alpha=0.05, 0.01, 0.001$ according to two-sided Mann-Whitney U-tests.


\textbf{H1}: We can see from both Fig.~\ref{fig:rst_noreact} and Fig.~\ref{fig:rst_react} that G-MPC achieves statistically significantly larger minimum distances to pedestrians across all scenarios, often with $p<0.001$. This illustrates that the group representation is in itself capable of upgrading a simple MPC with no prediction. As expected, we observe that the price G-MPC pays for that is a larger average path length. We also see that success rates are comparable. Overall, we conclude that H1 holds.

\textbf{H2}: When future state predictions are considered, G-MPC obtains statistically significant results in most scenes supporting its attributes of being safer at the cost of worse efficiency. Thus H2 is partially confirmed. In offline scenarios, G-MPC has lower success rates in crossing scenarios. Upon closer inspection, most failure cases are due to timeouts from G-MPC's conservative behavior. However, in online scenarios where pedestrians react to the robot, G-MPC achieves high success rates. In real-world situations, to cross dense traffic, the robot needs to plan its actions with expectations of reactive pedestrians. Otherwise, the robot will most probably run into \textit{the freezing robot problem}~\citep{trautmanijrr}. 

\textbf{H3}: Group-based representations have the potential to robustly account for imperfect state-estimates. Overall, we observe that with simulated imperfect states, G-MPC does not perform statistically significantly worse in terms of safety, but in dense crowds of the UNIV scenes it has worse efficiency and worse success rates in online cases. This shows that H3 holds in terms of safety and, in moderately dense human crowds, holds in terms of efficiency. Future work on better group representation is needed to achieve better efficiency in high-density human crowds given imperfect states.

\textbf{H4}: From Fig.~\ref{fig:rst_noreact} and Fig.~\ref{fig:rst_react}, we can see that G-MPC often has fewer group-space intrusions than its baselines. While this relationship is not always statistically significant, we do see a general trend of the group-based approaches to respect group spaces more often than individual ones. Thus, we conclude that H4 is partially confirmed.

We additionally observe a general trend that \textbf{group-pred} is better than \textbf{group-nopred} in terms of higher success rates, lower chances of group intrusions, longer minimum distances to pedestrians and shorter path lengths. This shows that our group prediction model offers benefits to the robot's navigation. However, in a few scenarios \textbf{group-nopred} performs better. We largely attribute this to the finite inaccuracies of future group predictions and the freezing robot problem that accompanies the robot's more conservative behavior in \textbf{group-pred} than in \textbf{group-nopred}.


\textbf{Qualitative Analysis.} Qualitatively, it is a more common occurrence for regular MPC to perform aggressive and socially inappropriate maneuvers than G-MPC. As shown in the two examples in Fig.~\ref{fig:rst_examples} executed by \textbf{ped-sgan} and \textbf{group-pred} agents, we can see that in offline conditions, the MPC agent aggressively cuts in front of the two pedestrians to the left before proceeding headlong into the cluster of pedestrians, only managing to avoid the deadlock by escaping through the narrow gap that opens up. While for G-MPC, it tracks the movements of the two pedestrian groups coming from the left. When the two pedestrian groups merge, the agent turns around and reevaluates its approach to cross. In the online condition, we observe that the MPC agent cuts through a pedestrian group to reach the other side, forcing a member of the group to stop and yield as indicated by the pedestrian's shrinking personal space, which is proportional to its speed. In the same situation, the G-MPC agent chooses to circumvent behind the social group.

%% file: sections/discussion.tex
\section{Conclusion}

We introduced a methodology for generating group-based representations and predicting their future states. Through an extensive evaluation over the flow and crossing scenarios drawn from 10 different real-world scenes from 2 different human datasets with both reactive and non-reactive agents, we demonstrate the value of group-based prediction in enabling safe and socially compliant navigation. Through experimentation with simulated laser scans, our model displays promising potential to process noisy sensor inputs without much performance downgrade.

Several improvements to our framework are possible. For example, we could incorporate state-of-the-art oracles in the form of advanced video prediction models~\citep{Guen-2020} or incorporate inter-group interaction modeling. Additional considerations such as the set of rollouts or the cost functions could possibly increase performance. We could also integrate our prediction model into alternative control frameworks such as reinforcement learning policies. 

Finally, we plan on validating our findings on a real-world robot to fully test the capability of G-MPC to handle noisy sensor inputs. We also plan to investigate ways to improve computation time to enhance our approach's real-world applicability. These include simplifying group representation geometry and predicting future group states in metric space instead of in image space.






%% file: sections/acknowledgment.tex
\section*{Acknowledgment}
This work was funded by grant (IIS-1734361) from the National Science Foundation and Honda Research Institute USA.

%% file: sections/appendix.tex
\newpage
\section*{Appendix}
\renewcommand{\thesubsection}{\Alph{subsection}}\label{sec:appendix}

\subsection{Personal Space Definition}\label{sec:ps_defn}

\begin{figure}[th!]
\centering
\includegraphics[width=0.3\linewidth]{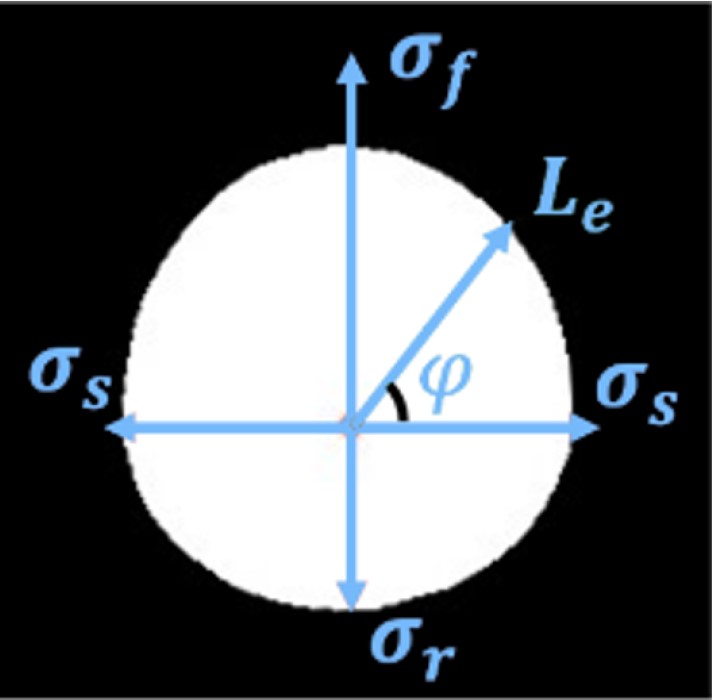}
\caption{A sample personal space}
\label{fig:PS}
\end{figure}

An example of the personal space is shown in Fig.~\ref{fig:PS}. Each personal space is first constructed by identifying the variances along the four principle axes to the agent's front, sides and rear, defined respectively as:
\begin{equation}
\begin{split}
    \sigma_{f}^i &= \max(2v^i, 0.5),\\ \sigma_{s}^i &=2\sigma_{f}^i/3\\ \sigma_{r}^i &=\sigma_{f}^i/2\\
\end{split}\mbox{.}
\end{equation}
Based on the four principle axis, the personal space for agent $i$ is represented as a set of boundary points.
\begin{equation}
    \mathcal{P}^i(q^i) = \{b^i(\phi), \phi \in [0,2\pi)\}\mbox{,}
\end{equation}
where
\begin{equation}
\begin{split}
    b^i(\phi) &= s^i + L_{e}(\phi)\left(\begin{matrix}
                        \cos(\theta^i+\phi) \\
                        \sin(\theta^i+\phi)
                   \end{matrix}\right)\mbox{,} \\
    L_e(\phi) &= \sqrt{\frac{C}{\cos^2{\gamma}/(2\sigma_1)+\sin^2{\gamma}/(2\sigma_2)}}\mbox{,} \\
    \gamma &= \mbox{mod}(\phi, \pi/2)\mbox{,} \\
    (\sigma_1, \sigma_2) &= \begin{cases}
        (\sigma_{f}^i, \sigma_{s}^i), &\mbox{if } 0 \leq \phi < \pi/2\\
        (\sigma_{s}^i, \sigma_{r}^i), &\mbox{if } \pi/2 \leq \phi < \pi\\
        (\sigma_{r}^i, \sigma_{s}^i), &\mbox{if } \pi \leq \phi < 3\pi/2\\
        (\sigma_{s}^i, \sigma_{f}^i), &\mbox{if } 3\pi/2 \leq \phi < 2\pi
        \end{cases}\mbox{.}
\end{split}
\label{eq:personal_space}
\end{equation}
$C$ is a constant used to adjust the scale of the personal space.

\subsection{Partial Input Handling}\label{sec:partial_input}

Note that in a dynamic pedestrian scene, we will have frequent occurrences of partial inputs for individual agents or groups due to new agents entering the scene or new groups being formed respectively. Therefore, our prediction model must be able to handle cases in which the input is complete up to a past window $t_{\hat{h}}$ with $t_{\hat{h}} = t -\hat{h}$, $\hat{h} < h$, i.e., $\boldsymbol{\mathcal{G}}_{t_{\hat{h}}:t}$. To handle these cases, for time $t_{h} <\tau < t_{\hat{h}}$, we compute $\mathcal{G}^j_{\tau}$ by making the following membership assumptions:
\begin{itemize}
    \item For any agent $i\in G^j_t$ such that $i\notin G^j_{\tau}$ and for whom we have the complete state history $s^i_{t_h:t}$, we set $g^i_{\tau} = j$. In other words, the prior group membership of any recent members of group $j$ is set to $j$ (despite agent $i$ possibly being a member of another group $j'$ at those instances). 
    
    \item For any agent $i\in G^j_t$ such that $i\notin G^j_{\tau}$ and for whom we only have \emph{partial} state history $s^i_{t_{\hat{h}}:t}$, we take the agent's last known state $s^{i}_t$ and velocity $u^{i}_t$ and back propagate it as $s_{\tau-1}^{i}=s_{\tau}^{i}-u_{\tau}^{i}dt$.
    
    

\end{itemize}
Given a small $h$, these assumptions should reflect a close approximation of the group's complete history state, because pedestrian group switching process is gradual and pedestrian movements are smooth and predictive. 

\subsection{Prediction Model Details}
\begin{figure}
    \centering
    \includegraphics[width=\linewidth]{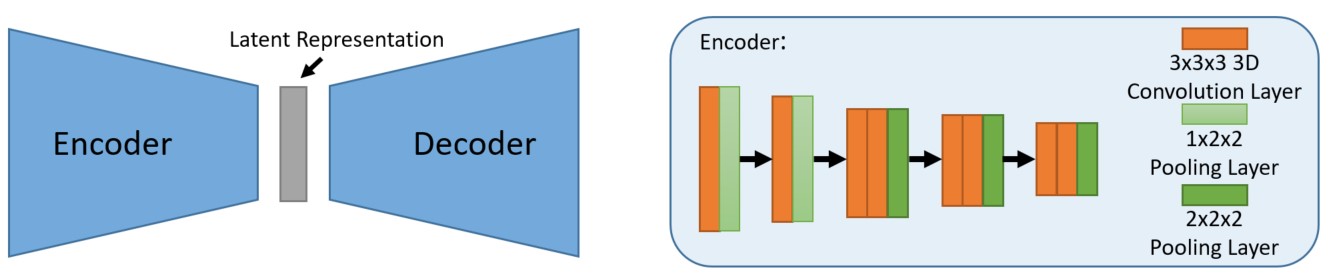}
    \caption{Our simple encoder-decoder network's architecture. The decoder's deconvolution layers mirror the layout of the encoder.}
    \label{fig:autoencoder}
\end{figure}

Our encoder-decoder network largely leverages [50]'s C3D network. As shown in Fig.~\ref{fig:autoencoder}, the encoder architecture contains the following layers (beginning from the input layer): one $3\times 3\times 3$ convolution layer with $64$ channels, one $1\times 2\times 2$ maxpool layer, one $3\times 3\times 3$ convolution layer with $128$ channels, another $1\times 2\times 2$ maxpool layer, another $3\times 3\times 3$ convolution layer with $128$ channels, one $3\times 3\times 3$ convolution layer with $256$ channels, one $2\times 2\times 2$ maxpool layer, another $3\times 3\times 3$ convolution layer with $256$ channels, one $3\times 3\times 3$ convolution layer with $512$ channels, another $2\times 2\times 2$ maxpool layer, two $3\times 3\times 3$ convolution layers with $512$ channels and another $2\times 2\times 2$ maxpool layer.

\begin{figure}
    \centering
    \includegraphics[width=\linewidth]{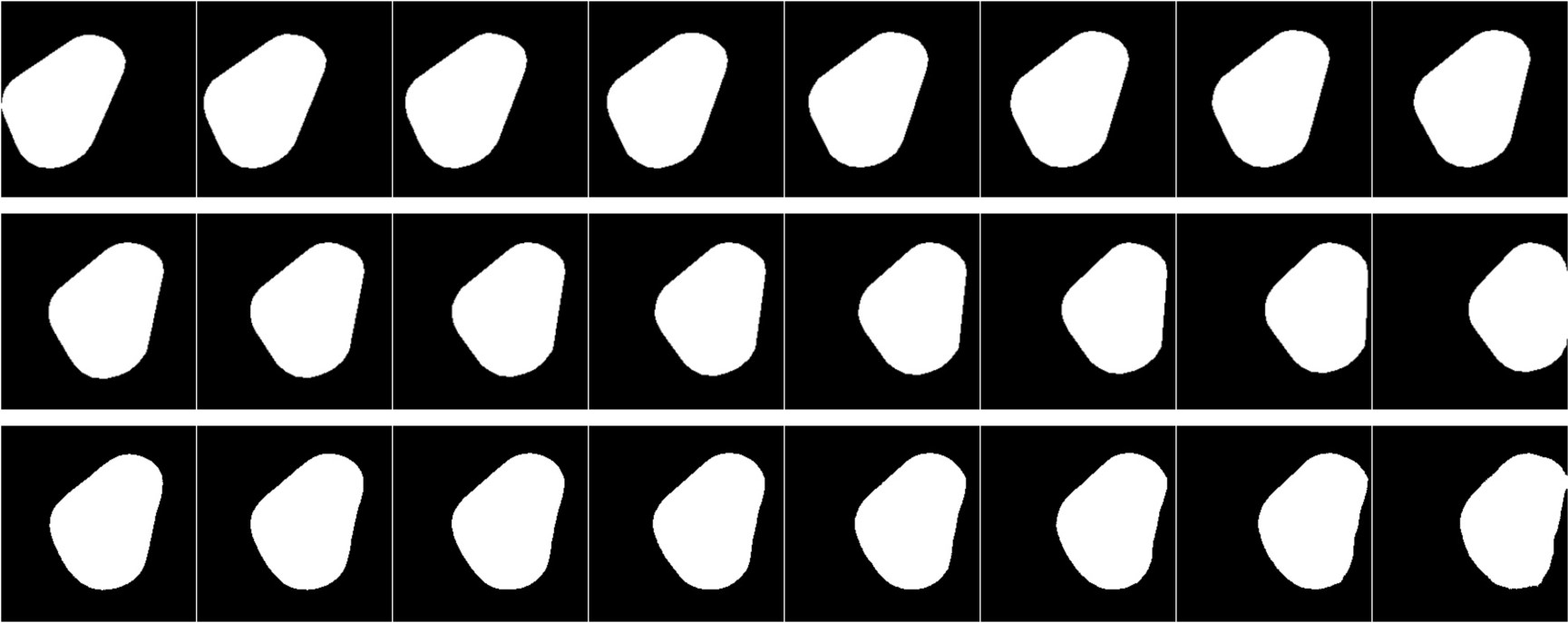}
    \caption{Top: An example group space input sequence for our encoder-decoder network. Mid: The ground truth future sequence of the group. Bottom: The predictioned future sequence of the group as outputted by our encoder-decoder network.}
    \label{fig:data_sample}
\end{figure}

We used an initial learning rate of $1e-5$, batch size of $1$ and trained for $200$ epochs. We used Adam optimizer with default PyTorch settings. The data samples are generated by sampling a random segment during the evolution of a group for all groups in all the datasets. The data samples are normalized in scale and positions such that the entire group space sequence fits inside the $224\times 224$ image sequences and the geometric center of the group in the last input sequence frame is at the center of the image. After obtaining the predictions from the model, we filter out pixel predictions with confidence level less than $0.5$. An example is shown in Fig.~\ref{fig:data_sample}. For evaluation on a particular dataset, including both evaluation of the encoder-decoder network's performance and the policies in the navigation setting, we use the model that was trained on the other four datasets.

In simulated laser scan settings, we do not retrain the group shape prediction models. Instead, we transfer the learned group shape prediction models on perfect perception settings directly into this new setting. We use a nearest neighbor approach based on geometric centers to identify the history sequence of a group in order to predict the group's future states. If the nearest neighbor of a group in the previous frame is more than $0.25m$ away, then we say no prior history of this group is available and use the technique in section~\ref{sec:partial_input} to linearly back-propagate the group's history.

To integrate this encoder-decoder network into our G-MPC framework, we performed the following additional processing steps (a detailed explanation of footnote 1). Because the encoder-decoder network model takes image-based inputs, we first convert the group space convex hulls into images. We use the homography matrix provided by the dataset to convert the coordinates of the convexhull vertices from meters to pixels. Then we draw these convexhulls on empty canvases to obtain the images. We preprocessed these images to normalize the convexhulls' sizes and positions such that these convexhulls fit inside the images throughout the input sequence and are approximately at the center of the image for the last input frame. Once we obtain the output sequence, we take the coordinates of the vertices at the edge of the output shape blobs and convert them back from pixels to meters using the inverse of the homography matrix.

\subsection{Parameter Details}

For the parameters of eq. (1), we picked $\epsilon_s, \epsilon_\theta, \epsilon_v$ such that the grouping outcomes match our qualitative inspection of human grouping in the datasets similarly to our prior work [13]. For ETH, HOTEL, ZARA1 and ZARA2 we set $\epsilon_s = 2.0m, \epsilon_\theta = 30^\circ, \epsilon_v=1.0 m/s$. Because UNIV is more crowded than the other four datasets, group formations are tighter and we set $\epsilon_s = 1.5m, \epsilon_\theta=15^\circ, \epsilon_v=0.5m/s$.
    
For the parameter of eq.~\eqref{eq:personal_space}, we selected $C$ under the assumption that closely-interacting pedestrians walk around the boundaries of each other's personal space. For ETH, HOTEL, ZARA1 and ZARA2, we set $C=0.35$. Again because UNIV has denser crowds, we set $C=0.25$. If at any given time the robot enters a social group space, we incrementally reduce $C$ by $0.1$ with a minimum value of $0.05$ until the robot is outside of the group space.
    
For the time horizon parameter $f$ and the history window parameter $h$ from section 4.2, we set $f=8$ and $h=8$ to ensure our MPC formulation's compatibility with the SGAN models.
    
For the weight parameter $\lambda$ in the cost function in equation (10), we perform a full parameter sweep to tune $\lambda$. We test $\lambda$ with values from $0.1$ to $0.9$ with increments of $0.05$ on randomly sampled $100$ test cases. We then select $\lambda$ that results in high success rates (at least $90\%$) for both agent based and group based policies without predictions and that the success rates of the two policies are the closest to each other. For trials with non-reactive agents, we set $\lambda = 0.65$. For trials with reactive agents, we set $\lambda = 0.3$. Note that we want the weight parameter to be the same for both pedestrian-based and group-based policies because the distance from the pedestrians to the boundaries of the social space are the same in both settings. Keeping the same weight allows fair evaluations of these two types of policies.
    
For the number of control rollouts $R$ in equation (14), we set $R = 12$.

\subsection{Numeric Results of Fig. 4 and Fig. 5}\label{sec:data}

\begin{table*}
\centering
\caption{Performance per scene under the \textit{Offline} condition.}
\resizebox{\columnwidth}{!}{%
\addvbuffer[0pt 2pt]{
\begin{tabular}{lccccccccccc}
\toprule
\multicolumn{1}{l}{Scene} &
\multicolumn{1}{l}{} &
 \multicolumn{2}{c}{ETH} & \multicolumn{2}{c}{HOTEL} & \multicolumn{2}{c}{ZARA1} &
\multicolumn{2}{c}{ZARA2} &
\multicolumn{2}{c}{UNIV}\\ 
\midrule

\multicolumn{1}{l|}{Task} &
\multicolumn{1}{l|}{Metric} &
\multicolumn{1}{c|}{\textbf{Flow}} & \multicolumn{1}{c|}{\textbf{Cross}} &
\multicolumn{1}{c|}{\textbf{Flow}} & \multicolumn{1}{c|}{\textbf{Cross}} &
\multicolumn{1}{c|}{\textbf{Flow}} & \multicolumn{1}{c|}{\textbf{Cross}} &
\multicolumn{1}{c|}{\textbf{Flow}} & \multicolumn{1}{c|}{\textbf{Cross}} &
\multicolumn{1}{c|}{\textbf{Flow}} & \multicolumn{1}{c|}{\textbf{Cross}}\\ 


\midrule
\multicolumn{1}{l|}{\multirow{5}{*}{\textbf{ped-nopred}}} &

\multicolumn{1}{c|}{$\mathcal{S} (\%)$} & 
\multicolumn{1}{c|}{$91.38$} & 
\multicolumn{1}{c|}{$75.86$} &
\multicolumn{1}{c|}{$95.35$} &
\multicolumn{1}{c|}{$95.45$} &
\multicolumn{1}{c|}{$96.00$} &
\multicolumn{1}{c|}{$85.71$} &
\multicolumn{1}{c|}{$97.64$} &
\multicolumn{1}{c|}{$96.90$} &
\multicolumn{1}{c|}{$83.96$} &
\multicolumn{1}{c|}{$83.33$}\\

\multicolumn{1}{c|}{} & \multicolumn{1}{c|}{$\mathcal{C} (\%)$} & 
\multicolumn{1}{c|}{$56.9$} & 
\multicolumn{1}{c|}{$24.14$} &
\multicolumn{1}{c|}{$81.4$} &
\multicolumn{1}{c|}{$63.64$} &
\multicolumn{1}{c|}{$72.0$} &
\multicolumn{1}{c|}{$46.43$} &
\multicolumn{1}{c|}{$82.68$} &
\multicolumn{1}{c|}{$70.54$} &
\multicolumn{1}{c|}{$68.87$} &
\multicolumn{1}{c|}{$64.91$}\\

\multicolumn{1}{c|}{} & \multicolumn{1}{c|}{$\mathcal{D} (m)$} & 
\multicolumn{1}{c|}{$1.29$} & 
\multicolumn{1}{c|}{$0.97$} &
\multicolumn{1}{c|}{$1.35$} &
\multicolumn{1}{c|}{$1.14$} &
\multicolumn{1}{c|}{$1.31$} &
\multicolumn{1}{c|}{$1.03$} &
\multicolumn{1}{c|}{$1.39$} &
\multicolumn{1}{c|}{$1.27$} &
\multicolumn{1}{c|}{$0.95$} &
\multicolumn{1}{c|}{$0.98$}\\

\multicolumn{1}{c|}{} & \multicolumn{1}{c|}{$\mathcal{L} (m)$} & 
\multicolumn{1}{c|}{$27.16$} & 
\multicolumn{1}{c|}{$16.93$} &
\multicolumn{1}{c|}{$19.31$} &
\multicolumn{1}{c|}{$10.55$} &
\multicolumn{1}{c|}{$20.75$} &
\multicolumn{1}{c|}{$8.80$} &
\multicolumn{1}{c|}{$20.87$} &
\multicolumn{1}{c|}{$8.92$} &
\multicolumn{1}{c|}{$22.96$} &
\multicolumn{1}{c|}{$17.38$}\\ \midrule

\multicolumn{1}{l|}{\multirow{5}{*}{\textbf{ped-linear}}} &

\multicolumn{1}{c|}{$\mathcal{S} (\%)$} & 
\multicolumn{1}{c|}{$94.83$} & 
\multicolumn{1}{c|}{$96.55$} &
\multicolumn{1}{c|}{$97.67$} &
\multicolumn{1}{c|}{$100$} &
\multicolumn{1}{c|}{$96$} &
\multicolumn{1}{c|}{$96.43$} &
\multicolumn{1}{c|}{$94.49$} &
\multicolumn{1}{c|}{$100$} &
\multicolumn{1}{c|}{$86.79$} &
\multicolumn{1}{c|}{$99.12$}\\

\multicolumn{1}{c|}{} & \multicolumn{1}{c|}{$\mathcal{C} (\%)$} & 
\multicolumn{1}{c|}{$75.86$} & 
\multicolumn{1}{c|}{$74.14$} &
\multicolumn{1}{c|}{$86.05$} &
\multicolumn{1}{c|}{$88.64$} &
\multicolumn{1}{c|}{$84.0$} &
\multicolumn{1}{c|}{$89.29$} &
\multicolumn{1}{c|}{$88.19$} &
\multicolumn{1}{c|}{$95.35$} &
\multicolumn{1}{c|}{$79.25$} &
\multicolumn{1}{c|}{$90.35$}\\

\multicolumn{1}{c|}{} & \multicolumn{1}{c|}{$\mathcal{D} (m)$} & 
\multicolumn{1}{c|}{$1.42$} & 
\multicolumn{1}{c|}{$1.28$} &
\multicolumn{1}{c|}{$1.28$} &
\multicolumn{1}{c|}{$1.25$} &
\multicolumn{1}{c|}{$1.53$} &
\multicolumn{1}{c|}{$1.36$} &
\multicolumn{1}{c|}{$1.53$} &
\multicolumn{1}{c|}{$1.46$} &
\multicolumn{1}{c|}{$1.01$} &
\multicolumn{1}{c|}{$1.09$}\\

\multicolumn{1}{c|}{} & \multicolumn{1}{c|}{$\mathcal{L} (m)$} & 
\multicolumn{1}{c|}{$27.31$} & 
\multicolumn{1}{c|}{$18.08$} &
\multicolumn{1}{c|}{$19.28$} &
\multicolumn{1}{c|}{$10.72$} &
\multicolumn{1}{c|}{$20.29$} &
\multicolumn{1}{c|}{$9.48$} &
\multicolumn{1}{c|}{$20.84$} &
\multicolumn{1}{c|}{$9.16$} &
\multicolumn{1}{c|}{$22.93$} &
\multicolumn{1}{c|}{$17.06$}\\


\midrule
\multicolumn{1}{l|}{\multirow{5}{*}{\textbf{ped-sgan}}} &

\multicolumn{1}{c|}{$\mathcal{S} (\%)$} & 
\multicolumn{1}{c|}{$93.1$} & 
\multicolumn{1}{c|}{$94.83$} &
\multicolumn{1}{c|}{$95.35$} &
\multicolumn{1}{c|}{$100$} &
\multicolumn{1}{c|}{$96$} &
\multicolumn{1}{c|}{$100$} &
\multicolumn{1}{c|}{$96.06$} &
\multicolumn{1}{c|}{$100$} &
\multicolumn{1}{c|}{$86.79$} &
\multicolumn{1}{c|}{$100$}\\

\multicolumn{1}{c|}{} & \multicolumn{1}{c|}{$\mathcal{C} (\%)$} & 
\multicolumn{1}{c|}{$79.31$} & 
\multicolumn{1}{c|}{$70.69$} &
\multicolumn{1}{c|}{$81.4$} &
\multicolumn{1}{c|}{$79.55$} &
\multicolumn{1}{c|}{$80.0$} &
\multicolumn{1}{c|}{$92.86$} &
\multicolumn{1}{c|}{$89.76$} &
\multicolumn{1}{c|}{$95.35$} &
\multicolumn{1}{c|}{$73.58$} &
\multicolumn{1}{c|}{$86.84$}\\

\multicolumn{1}{c|}{} & \multicolumn{1}{c|}{$\mathcal{D} (m)$} & 
\multicolumn{1}{c|}{$1.45$} & 
\multicolumn{1}{c|}{$1.27$} &
\multicolumn{1}{c|}{$1.34$} &
\multicolumn{1}{c|}{$1.23$} &
\multicolumn{1}{c|}{$1.46$} &
\multicolumn{1}{c|}{$1.35$} &
\multicolumn{1}{c|}{$1.50$} &
\multicolumn{1}{c|}{$1.47$} &
\multicolumn{1}{c|}{$0.98$} &
\multicolumn{1}{c|}{$1.08$}\\

\multicolumn{1}{c|}{} & \multicolumn{1}{c|}{$\mathcal{L} (m)$} & 
\multicolumn{1}{c|}{$27.05$} & 
\multicolumn{1}{c|}{$17.99$} &
\multicolumn{1}{c|}{$19.20$} &
\multicolumn{1}{c|}{$10.10$} &
\multicolumn{1}{c|}{$20.51$} &
\multicolumn{1}{c|}{$9.66$} &
\multicolumn{1}{c|}{$20.83$} &
\multicolumn{1}{c|}{$9.21$} &
\multicolumn{1}{c|}{$23.05$} &
\multicolumn{1}{c|}{$17.22$}\\


\midrule
\multicolumn{1}{l|}{\multirow{5}{*}{\textbf{group-nopred}}} &

\multicolumn{1}{c|}{$\mathcal{S} (\%)$} & 
\multicolumn{1}{c|}{$94.83$} & 
\multicolumn{1}{c|}{$87.93$} &
\multicolumn{1}{c|}{$100$} &
\multicolumn{1}{c|}{$88.64$} &
\multicolumn{1}{c|}{$92$} &
\multicolumn{1}{c|}{$75$} &
\multicolumn{1}{c|}{$96.06$} &
\multicolumn{1}{c|}{$92.25$} &
\multicolumn{1}{c|}{$93.4$} &
\multicolumn{1}{c|}{$93.86$}\\

\multicolumn{1}{c|}{} & \multicolumn{1}{c|}{$\mathcal{C} (\%)$} & 
\multicolumn{1}{c|}{$82.76$} & 
\multicolumn{1}{c|}{$79.31$} &
\multicolumn{1}{c|}{$97.67$} &
\multicolumn{1}{c|}{$86.36$} &
\multicolumn{1}{c|}{$88.0$} &
\multicolumn{1}{c|}{$71.43$} &
\multicolumn{1}{c|}{$92.91$} &
\multicolumn{1}{c|}{$89.92$} &
\multicolumn{1}{c|}{$90.57$} &
\multicolumn{1}{c|}{$88.6$}\\

\multicolumn{1}{c|}{} & \multicolumn{1}{c|}{$\mathcal{D} (m)$} & 
\multicolumn{1}{c|}{$1.70$} & 
\multicolumn{1}{c|}{$1.61$} &
\multicolumn{1}{c|}{$1.8$} &
\multicolumn{1}{c|}{$1.61$} &
\multicolumn{1}{c|}{$1.83$} &
\multicolumn{1}{c|}{$1.59$} &
\multicolumn{1}{c|}{$1.68$} &
\multicolumn{1}{c|}{$1.59$} &
\multicolumn{1}{c|}{$1.18$} &
\multicolumn{1}{c|}{$1.22$}\\

\multicolumn{1}{c|}{} & \multicolumn{1}{c|}{$\mathcal{L} (m)$} & 
\multicolumn{1}{c|}{$29.52$} & 
\multicolumn{1}{c|}{$23.32$} &
\multicolumn{1}{c|}{$21.98$} &
\multicolumn{1}{c|}{$14.38$} &
\multicolumn{1}{c|}{$22.86$} &
\multicolumn{1}{c|}{$13.02$} &
\multicolumn{1}{c|}{$23.47$} &
\multicolumn{1}{c|}{$11.17$} &
\multicolumn{1}{c|}{$28.57$} &
\multicolumn{1}{c|}{$20.61$}\\


\midrule
\multicolumn{1}{l|}{\multirow{5}{*}{\textbf{group-pred}}} &

\multicolumn{1}{c|}{$\mathcal{S} (\%)$} & 
\multicolumn{1}{c|}{$96.55$} & 
\multicolumn{1}{c|}{$86.21$} &
\multicolumn{1}{c|}{$100$} &
\multicolumn{1}{c|}{$90.91$} &
\multicolumn{1}{c|}{$96$} &
\multicolumn{1}{c|}{$82.14$} &
\multicolumn{1}{c|}{$96.85$} &
\multicolumn{1}{c|}{$94.57$} &
\multicolumn{1}{c|}{$89.62$} &
\multicolumn{1}{c|}{$93.86$}\\

\multicolumn{1}{c|}{} & \multicolumn{1}{c|}{$\mathcal{C} (\%)$} & 
\multicolumn{1}{c|}{$82.76$} & 
\multicolumn{1}{c|}{$81.03$} &
\multicolumn{1}{c|}{$100.0$} &
\multicolumn{1}{c|}{$88.64$} &
\multicolumn{1}{c|}{$92.0$} &
\multicolumn{1}{c|}{$82.14$} &
\multicolumn{1}{c|}{$92.91$} &
\multicolumn{1}{c|}{$93.02$} &
\multicolumn{1}{c|}{$86.79$} &
\multicolumn{1}{c|}{$92.98$}\\

\multicolumn{1}{c|}{} & \multicolumn{1}{c|}{$\mathcal{D} (m)$} & 
\multicolumn{1}{c|}{$1.67$} & 
\multicolumn{1}{c|}{$1.90$} &
\multicolumn{1}{c|}{$1.83$} &
\multicolumn{1}{c|}{$1.65$} &
\multicolumn{1}{c|}{$1.87$} &
\multicolumn{1}{c|}{$1.75$} &
\multicolumn{1}{c|}{$1.77$} &
\multicolumn{1}{c|}{$1.67$} &
\multicolumn{1}{c|}{$1.19$} &
\multicolumn{1}{c|}{$1.32$}\\

\multicolumn{1}{c|}{} & \multicolumn{1}{c|}{$\mathcal{L} (m)$} & 
\multicolumn{1}{c|}{$29.51$} & 
\multicolumn{1}{c|}{$23.17$} &
\multicolumn{1}{c|}{$21.88$} &
\multicolumn{1}{c|}{$13.63$} &
\multicolumn{1}{c|}{$23.01$} &
\multicolumn{1}{c|}{$11.95$} &
\multicolumn{1}{c|}{$23.45$} &
\multicolumn{1}{c|}{$11.13$} &
\multicolumn{1}{c|}{$27.82$} &
\multicolumn{1}{c|}{$20.06$}\\


\midrule
\multicolumn{1}{l|}{\multirow{5}{*}{\textbf{laser-group-pred}}} &

\multicolumn{1}{c|}{$\mathcal{S} (\%)$} & 
\multicolumn{1}{c|}{$94.83$} & 
\multicolumn{1}{c|}{$84.48$} &
\multicolumn{1}{c|}{$97.67$} &
\multicolumn{1}{c|}{$95.54$} &
\multicolumn{1}{c|}{$96$} &
\multicolumn{1}{c|}{$75$} &
\multicolumn{1}{c|}{$93.7$} &
\multicolumn{1}{c|}{$86.05$} &
\multicolumn{1}{c|}{$83.02$} &
\multicolumn{1}{c|}{$89.47$}\\

\multicolumn{1}{c|}{} & \multicolumn{1}{c|}{$\mathcal{C} (\%)$} & 
\multicolumn{1}{c|}{$81.03$} & 
\multicolumn{1}{c|}{$82.76$} &
\multicolumn{1}{c|}{$97.67$} &
\multicolumn{1}{c|}{$90.91$} &
\multicolumn{1}{c|}{$92.0$} &
\multicolumn{1}{c|}{$75.0$} &
\multicolumn{1}{c|}{$92.13$} &
\multicolumn{1}{c|}{$86.05$} &
\multicolumn{1}{c|}{$82.08$} &
\multicolumn{1}{c|}{$85.09$}\\

\multicolumn{1}{c|}{} & \multicolumn{1}{c|}{$\mathcal{D} (m)$} & 
\multicolumn{1}{c|}{$1.75$} & 
\multicolumn{1}{c|}{$2.21$} &
\multicolumn{1}{c|}{$1.86$} &
\multicolumn{1}{c|}{$1.70$} &
\multicolumn{1}{c|}{$1.92$} &
\multicolumn{1}{c|}{$1.81$} &
\multicolumn{1}{c|}{$1.8$} &
\multicolumn{1}{c|}{$1.76$} &
\multicolumn{1}{c|}{$1.25$} &
\multicolumn{1}{c|}{$1.40$}\\

\multicolumn{1}{c|}{} & \multicolumn{1}{c|}{$\mathcal{L} (m)$} & 
\multicolumn{1}{c|}{$29.57$} & 
\multicolumn{1}{c|}{$23.99$} &
\multicolumn{1}{c|}{$22.42$} &
\multicolumn{1}{c|}{$15.45$} &
\multicolumn{1}{c|}{$23.50$} &
\multicolumn{1}{c|}{$12.26$} &
\multicolumn{1}{c|}{$23.63$} &
\multicolumn{1}{c|}{$11.58$} &
\multicolumn{1}{c|}{$29.49$} &
\multicolumn{1}{c|}{$22.36$}\\

\bottomrule
\end{tabular}
}
}
\label{tab:results}
\end{table*}


\begin{table*}
\centering
\caption{Performance per scene under the \textit{Online} condition (simulated pedestrians powered by ORCA\cite{ORCA}).}
\resizebox{\columnwidth}{!}{%
\addvbuffer[0pt 2pt]{
\begin{tabular}{lccccccccccc}
\toprule
\multicolumn{1}{l}{Scene} &
\multicolumn{1}{l}{} &
 \multicolumn{2}{c}{ETH} & \multicolumn{2}{c}{HOTEL} & \multicolumn{2}{c}{ZARA1} &
\multicolumn{2}{c}{ZARA2} &
\multicolumn{2}{c}{UNIV}\\ 
\midrule

\multicolumn{1}{l|}{Task} &
\multicolumn{1}{l|}{Metric} &
\multicolumn{1}{c|}{\textbf{Flow}} & \multicolumn{1}{c|}{\textbf{Cross}} &
\multicolumn{1}{c|}{\textbf{Flow}} & \multicolumn{1}{c|}{\textbf{Cross}} &
\multicolumn{1}{c|}{\textbf{Flow}} & \multicolumn{1}{c|}{\textbf{Cross}} &
\multicolumn{1}{c|}{\textbf{Flow}} & \multicolumn{1}{c|}{\textbf{Cross}} &
\multicolumn{1}{c|}{\textbf{Flow}} & \multicolumn{1}{c|}{\textbf{Cross}}\\ 


\midrule
\multicolumn{1}{l|}{\multirow{5}{*}{\textbf{ped-nopred}}} &

\multicolumn{1}{c|}{$\mathcal{S} (\%)$} & 
\multicolumn{1}{c|}{$96.55$} & 
\multicolumn{1}{c|}{$98.28$} &
\multicolumn{1}{c|}{$100$} &
\multicolumn{1}{c|}{$97.73$} &
\multicolumn{1}{c|}{$100$} &
\multicolumn{1}{c|}{$100$} &
\multicolumn{1}{c|}{$97.64$} &
\multicolumn{1}{c|}{$100$} &
\multicolumn{1}{c|}{$88.68$} &
\multicolumn{1}{c|}{$99.12$}\\

\multicolumn{1}{c|}{} & \multicolumn{1}{c|}{$\mathcal{C} (\%)$} & 
\multicolumn{1}{c|}{$50.0$} & 
\multicolumn{1}{c|}{$63.79$} &
\multicolumn{1}{c|}{$79.07$} &
\multicolumn{1}{c|}{$77.27$} &
\multicolumn{1}{c|}{$60.0$} &
\multicolumn{1}{c|}{$64.29$} &
\multicolumn{1}{c|}{$65.35$} &
\multicolumn{1}{c|}{$79.84$} &
\multicolumn{1}{c|}{$60.38$} &
\multicolumn{1}{c|}{$78.95$}\\

\multicolumn{1}{c|}{} & \multicolumn{1}{c|}{$\mathcal{D} (m)$} & 
\multicolumn{1}{c|}{$0.93$} & 
\multicolumn{1}{c|}{$1.05$} &
\multicolumn{1}{c|}{$0.94$} &
\multicolumn{1}{c|}{$1.04$} &
\multicolumn{1}{c|}{$0.92$} &
\multicolumn{1}{c|}{$0.97$} &
\multicolumn{1}{c|}{$0.98$} &
\multicolumn{1}{c|}{$0.99$} &
\multicolumn{1}{c|}{$0.89$} &
\multicolumn{1}{c|}{$0.98$}\\

\multicolumn{1}{c|}{} & \multicolumn{1}{c|}{$\mathcal{L} (m)$} & 
\multicolumn{1}{c|}{$24.02$} & 
\multicolumn{1}{c|}{$13.73$} &
\multicolumn{1}{c|}{$15.75$} &
\multicolumn{1}{c|}{$8.50$} &
\multicolumn{1}{c|}{$16.71$} &
\multicolumn{1}{c|}{$6.13$} &
\multicolumn{1}{c|}{$17.16$} &
\multicolumn{1}{c|}{$7.32$} &
\multicolumn{1}{c|}{$17.20$} &
\multicolumn{1}{c|}{$13.82$}\\ \midrule

\multicolumn{1}{l|}{\multirow{5}{*}{\textbf{ped-linear}}} &

\multicolumn{1}{c|}{$\mathcal{S} (\%)$} & 
\multicolumn{1}{c|}{$94.83$} & 
\multicolumn{1}{c|}{$98.28$} &
\multicolumn{1}{c|}{$100$} &
\multicolumn{1}{c|}{$97.73$} &
\multicolumn{1}{c|}{$100$} &
\multicolumn{1}{c|}{$100$} &
\multicolumn{1}{c|}{$98.43$} &
\multicolumn{1}{c|}{$100$} &
\multicolumn{1}{c|}{$89.62$} &
\multicolumn{1}{c|}{$99.12$}\\

\multicolumn{1}{c|}{} & \multicolumn{1}{c|}{$\mathcal{C} (\%)$} & 
\multicolumn{1}{c|}{$43.1$} & 
\multicolumn{1}{c|}{$70.69$} &
\multicolumn{1}{c|}{$83.72$} &
\multicolumn{1}{c|}{$90.91$} &
\multicolumn{1}{c|}{$64.0$} &
\multicolumn{1}{c|}{$85.71$} &
\multicolumn{1}{c|}{$74.8$} &
\multicolumn{1}{c|}{$80.62$} &
\multicolumn{1}{c|}{$61.32$} &
\multicolumn{1}{c|}{$88.6$}\\

\multicolumn{1}{c|}{} & \multicolumn{1}{c|}{$\mathcal{D} (m)$} & 
\multicolumn{1}{c|}{$0.94$} & 
\multicolumn{1}{c|}{$1.04$} &
\multicolumn{1}{c|}{$0.97$} &
\multicolumn{1}{c|}{$1.03$} &
\multicolumn{1}{c|}{$0.94$} &
\multicolumn{1}{c|}{$0.99$} &
\multicolumn{1}{c|}{$0.98$} &
\multicolumn{1}{c|}{$0.99$} &
\multicolumn{1}{c|}{$0.90$} &
\multicolumn{1}{c|}{$0.98$}\\

\multicolumn{1}{c|}{} & \multicolumn{1}{c|}{$\mathcal{L} (m)$} & 
\multicolumn{1}{c|}{$23.83$} & 
\multicolumn{1}{c|}{$13.25$} &
\multicolumn{1}{c|}{$15.43$} &
\multicolumn{1}{c|}{$8.32$} &
\multicolumn{1}{c|}{$16.54$} &
\multicolumn{1}{c|}{$6.22$} &
\multicolumn{1}{c|}{$17.03$} &
\multicolumn{1}{c|}{$6.74$} &
\multicolumn{1}{c|}{$16.87$} &
\multicolumn{1}{c|}{$13.53$}\\


\midrule
\multicolumn{1}{l|}{\multirow{5}{*}{\textbf{ped-sgan}}} &

\multicolumn{1}{c|}{$\mathcal{S} (\%)$} & 
\multicolumn{1}{c|}{$96.55$} & 
\multicolumn{1}{c|}{$98.28$} &
\multicolumn{1}{c|}{$100$} &
\multicolumn{1}{c|}{$97.73$} &
\multicolumn{1}{c|}{$100$} &
\multicolumn{1}{c|}{$100$} &
\multicolumn{1}{c|}{$98.43$} &
\multicolumn{1}{c|}{$100$} &
\multicolumn{1}{c|}{$89.62$} &
\multicolumn{1}{c|}{$99.12$}\\

\multicolumn{1}{c|}{} & \multicolumn{1}{c|}{$\mathcal{C} (\%)$} & 
\multicolumn{1}{c|}{$39.66$} & 
\multicolumn{1}{c|}{$63.79$} &
\multicolumn{1}{c|}{$88.37$} &
\multicolumn{1}{c|}{$88.64$} &
\multicolumn{1}{c|}{$64.0$} &
\multicolumn{1}{c|}{$82.14$} &
\multicolumn{1}{c|}{$77.95$} &
\multicolumn{1}{c|}{$82.95$} &
\multicolumn{1}{c|}{$62.26$} &
\multicolumn{1}{c|}{$86.84$}\\

\multicolumn{1}{c|}{} & \multicolumn{1}{c|}{$\mathcal{D} (m)$} & 
\multicolumn{1}{c|}{$0.93$} & 
\multicolumn{1}{c|}{$1.04$} &
\multicolumn{1}{c|}{$0.95$} &
\multicolumn{1}{c|}{$1.04$} &
\multicolumn{1}{c|}{$0.94$} &
\multicolumn{1}{c|}{$0.99$} &
\multicolumn{1}{c|}{$0.98$} &
\multicolumn{1}{c|}{$0.99$} &
\multicolumn{1}{c|}{$0.90$} &
\multicolumn{1}{c|}{$0.98$}\\

\multicolumn{1}{c|}{} & \multicolumn{1}{c|}{$\mathcal{L} (m)$} & 
\multicolumn{1}{c|}{$23.85$} & 
\multicolumn{1}{c|}{$13.20$} &
\multicolumn{1}{c|}{$15.63$} &
\multicolumn{1}{c|}{$8.14$} &
\multicolumn{1}{c|}{$16.54$} &
\multicolumn{1}{c|}{$6.18$} &
\multicolumn{1}{c|}{$17.06$} &
\multicolumn{1}{c|}{$6.72$} &
\multicolumn{1}{c|}{$16.90$} &
\multicolumn{1}{c|}{$13.53$}\\


\midrule
\multicolumn{1}{l|}{\multirow{5}{*}{\textbf{group-nopred}}} &

\multicolumn{1}{c|}{$\mathcal{S} (\%)$} & 
\multicolumn{1}{c|}{$93.1$} & 
\multicolumn{1}{c|}{$98.28$} &
\multicolumn{1}{c|}{$95.35$} &
\multicolumn{1}{c|}{$100$} &
\multicolumn{1}{c|}{$96$} &
\multicolumn{1}{c|}{$100$} &
\multicolumn{1}{c|}{$97.64$} &
\multicolumn{1}{c|}{$98.45$} &
\multicolumn{1}{c|}{$89.62$} &
\multicolumn{1}{c|}{$100$}\\

\multicolumn{1}{c|}{} & \multicolumn{1}{c|}{$\mathcal{C} (\%)$} & 
\multicolumn{1}{c|}{$75.86$} & 
\multicolumn{1}{c|}{$79.31$} &
\multicolumn{1}{c|}{$81.4$} &
\multicolumn{1}{c|}{$88.64$} &
\multicolumn{1}{c|}{$88.0$} &
\multicolumn{1}{c|}{$71.43$} &
\multicolumn{1}{c|}{$90.55$} &
\multicolumn{1}{c|}{$85.27$} &
\multicolumn{1}{c|}{$68.87$} &
\multicolumn{1}{c|}{$88.6$}\\

\multicolumn{1}{c|}{} & \multicolumn{1}{c|}{$\mathcal{D} (m)$} & 
\multicolumn{1}{c|}{$1.15$} & 
\multicolumn{1}{c|}{$1.15$} &
\multicolumn{1}{c|}{$1.17$} &
\multicolumn{1}{c|}{$1.16$} &
\multicolumn{1}{c|}{$1.11$} &
\multicolumn{1}{c|}{$1.03$} &
\multicolumn{1}{c|}{$1.21$} &
\multicolumn{1}{c|}{$1.11$} &
\multicolumn{1}{c|}{$0.94$} &
\multicolumn{1}{c|}{$1.04$}\\

\multicolumn{1}{c|}{} & \multicolumn{1}{c|}{$\mathcal{L} (m)$} & 
\multicolumn{1}{c|}{$25.36$} & 
\multicolumn{1}{c|}{$15.37$} &
\multicolumn{1}{c|}{$16.62$} &
\multicolumn{1}{c|}{$9.72$} &
\multicolumn{1}{c|}{$18.16$} &
\multicolumn{1}{c|}{$7.50$} &
\multicolumn{1}{c|}{$18.36$} &
\multicolumn{1}{c|}{$8.56$} &
\multicolumn{1}{c|}{$18.98$} &
\multicolumn{1}{c|}{$15.15$}\\


\midrule
\multicolumn{1}{l|}{\multirow{5}{*}{\textbf{group-pred}}} &

\multicolumn{1}{c|}{$\mathcal{S} (\%)$} & 
\multicolumn{1}{c|}{$94.83$} & 
\multicolumn{1}{c|}{$98.28$} &
\multicolumn{1}{c|}{$97.67$} &
\multicolumn{1}{c|}{$97.73$} &
\multicolumn{1}{c|}{$92$} &
\multicolumn{1}{c|}{$96.43$} &
\multicolumn{1}{c|}{$99.21$} &
\multicolumn{1}{c|}{$99.22$} &
\multicolumn{1}{c|}{$93.4$} &
\multicolumn{1}{c|}{$98.25$}\\

\multicolumn{1}{c|}{} & \multicolumn{1}{c|}{$\mathcal{C} (\%)$} & 
\multicolumn{1}{c|}{$74.14$} & 
\multicolumn{1}{c|}{$87.93$} &
\multicolumn{1}{c|}{$81.4$} &
\multicolumn{1}{c|}{$95.45$} &
\multicolumn{1}{c|}{$88.0$} &
\multicolumn{1}{c|}{$89.29$} &
\multicolumn{1}{c|}{$91.34$} &
\multicolumn{1}{c|}{$93.8$} &
\multicolumn{1}{c|}{$80.19$} &
\multicolumn{1}{c|}{$91.23$}\\

\multicolumn{1}{c|}{} & \multicolumn{1}{c|}{$\mathcal{D} (m)$} & 
\multicolumn{1}{c|}{$1.13$} & 
\multicolumn{1}{c|}{$1.24$} &
\multicolumn{1}{c|}{$1.16$} &
\multicolumn{1}{c|}{$1.21$} &
\multicolumn{1}{c|}{$1.17$} &
\multicolumn{1}{c|}{$1.13$} &
\multicolumn{1}{c|}{$1.18$} &
\multicolumn{1}{c|}{$1.11$} &
\multicolumn{1}{c|}{$0.93$} &
\multicolumn{1}{c|}{$1.07$}\\

\multicolumn{1}{c|}{} & \multicolumn{1}{c|}{$\mathcal{L} (m)$} & 
\multicolumn{1}{c|}{$25.19$} & 
\multicolumn{1}{c|}{$14.99$} &
\multicolumn{1}{c|}{$16.45$} &
\multicolumn{1}{c|}{$9.14$} &
\multicolumn{1}{c|}{$17.93$} &
\multicolumn{1}{c|}{$7.62$} &
\multicolumn{1}{c|}{$18.22$} &
\multicolumn{1}{c|}{$7.88$} &
\multicolumn{1}{c|}{$18.71$} &
\multicolumn{1}{c|}{$14.74$}\\


\midrule
\multicolumn{1}{l|}{\multirow{5}{*}{\textbf{laser-group-pred}}} &

\multicolumn{1}{c|}{$\mathcal{S} (\%)$} & 
\multicolumn{1}{c|}{$91.38$} & 
\multicolumn{1}{c|}{$98.28$} &
\multicolumn{1}{c|}{$95.35$} &
\multicolumn{1}{c|}{$97.73$} &
\multicolumn{1}{c|}{$96$} &
\multicolumn{1}{c|}{$89.29$} &
\multicolumn{1}{c|}{$97.64$} &
\multicolumn{1}{c|}{$97.67$} &
\multicolumn{1}{c|}{$84.91$} &
\multicolumn{1}{c|}{$92.11$}\\

\multicolumn{1}{c|}{} & \multicolumn{1}{c|}{$\mathcal{C} (\%)$} & 
\multicolumn{1}{c|}{$70.69$} & 
\multicolumn{1}{c|}{$84.48$} &
\multicolumn{1}{c|}{$88.37$} &
\multicolumn{1}{c|}{$93.18$} &
\multicolumn{1}{c|}{$88.0$} &
\multicolumn{1}{c|}{$85.71$} &
\multicolumn{1}{c|}{$88.98$} &
\multicolumn{1}{c|}{$94.57$} &
\multicolumn{1}{c|}{$60.38$} &
\multicolumn{1}{c|}{$78.07$}\\

\multicolumn{1}{c|}{} & \multicolumn{1}{c|}{$\mathcal{D} (m)$} & 
\multicolumn{1}{c|}{$1.18$} & 
\multicolumn{1}{c|}{$1.33$} &
\multicolumn{1}{c|}{$1.26$} &
\multicolumn{1}{c|}{$1.28$} &
\multicolumn{1}{c|}{$1.18$} &
\multicolumn{1}{c|}{$1.12$} &
\multicolumn{1}{c|}{$1.23$} &
\multicolumn{1}{c|}{$1.14$} &
\multicolumn{1}{c|}{$0.93$} &
\multicolumn{1}{c|}{$1.07$}\\

\multicolumn{1}{c|}{} & \multicolumn{1}{c|}{$\mathcal{L} (m)$} & 
\multicolumn{1}{c|}{$25.40$} & 
\multicolumn{1}{c|}{$16.27$} &
\multicolumn{1}{c|}{$16.81$} &
\multicolumn{1}{c|}{$9.82$} &
\multicolumn{1}{c|}{$18.72$} &
\multicolumn{1}{c|}{$8.54$} &
\multicolumn{1}{c|}{$19.07$} &
\multicolumn{1}{c|}{$8.57$} &
\multicolumn{1}{c|}{$21.39$} &
\multicolumn{1}{c|}{$16.46$}\\

\bottomrule
\end{tabular}
}
}
\label{tab:results-RVO}
\end{table*}

Tab.~\ref{tab:results} and Tab.~\ref{tab:results-RVO} are the numerical results of Fig. 4 and Fig. 5. $\mathcal{S}$ is the success rate. $\mathcal{C}$ is percentage of trials in which the robot does not enter any group space (collisions also count as group intrusions). $\mathcal{D}$ is the average minimum distance to pedestrians. $\mathcal{L}$ is the average path length.